\documentclass[sigconf]{acmart}

\AtBeginDocument{%
  }

\setcopyright{acmlicensed}
\copyrightyear{2018}
\acmYear{2018}
\acmDOI{XXXXXXX.XXXXXXX}

\usepackage{multirow}
\usepackage{subfigure}
\usepackage{graphicx}

\begin{document}

\begin{sloppypar}
\title{A Benchmark for Gaussian Splatting Compression and Quality Assessment Study}

\author{Qi Yang$^{\#}$*,\quad Kaifa Yang$^{\dagger}$*,\quad Yuke Xing$^{\dagger}$,\quad Yiling Xu$^{\dagger}$,\quad Zhu Li$^{\#}$}

\affiliation{
 \institution{
 $^{\#}$: School of Science and Engineering, University of Missouri - Kansas City}
  \city{}
  \country{}
 \institution{
 $^{\dagger}$: Cooperative Medianet Innovation Center, Shanghai Jiaotong University}
  \city{}
  \country{}
  }
  \email{littlleempty@gmail.com, {sekiroyyy, xingyuke-v, yl_xu}@sjtu.edu.cn, lizhu@umkc.edu}

  \thanks{*: Equal contribution}


\begin{abstract}
Due to the unconstrained densification and high-dimension primitive attributes, 3D Gaussian Splatting (GS) compression has evolved in conjunction with the work of 3D reconstruction and attracted considerable attention. Current GS compression studies focus mainly on how to realize a more compact scene representation, such as using fewer GS primitives or convert explicit GS data into implicit form, which can be classified as the ``generative compression'' method. However, the compression studies on the GS data itself, which can be regarded as ``traditional compression'', are empty. To fill the gap of traditional GS compression method, in this paper, we first propose a simple and effective GS data compression anchor called Graph-based GS Compression (GGSC). GGSC is inspired by graph signal processing theory and uses two branches to compress the primitive center and attributes. We split the whole GS sample via KDTree and clip the high-frequency components after the graph Fourier transform. Followed by quantization, G-PCC and adaptive arithmetic coding are used to compress the primitive center and attribute residual matrix to generate the bitrate file. GGSS is the first work to explore traditional GS compression, with advantages that can reveal the GS distortion characteristics corresponding to typical compression operation, such as high-frequency clipping and quantization. Second, based on GGSC, we create a GS Quality Assessment dataset (GSQA) with 120 samples. A subjective experiment is conducted in a laboratory environment to collect subjective scores after rendering GS into Processed Video Sequences (PVS). We analyze the characteristics of different GS distortions based on Mean Opinion Scores (MOS), demonstrating the sensitivity of different attributes distortion to visual quality. The GGSC code and the dataset, including GS samples, MOS, and PVS, are made publicly available at \url{https://github.com/Qi-Yangsjtu/GGSC}.
\end{abstract}



\maketitle



\section{Introduction}
Amazed by the impressive quality-complexity tradeoff in 3D scene representation, 3D Gaussian Splatting (GS) \cite{GS} has attracted considerable attention in both academic and industrial field. Different from Neural Radiance Field (NeRF) \cite{nerf} built on continuous scene representations, 3D GS uses serial scattered isotropic ellipsoids to reconstruct the 3D scene. Each primitive consists of several important attributes, including center, color Spherical Harmonic coefficients (SH), opacity, scale, and rotation. With high-efficiency customized CUDA-based rendering implementation, 3D GS can realize fast training and rendering, which provides great potential for practical applications.

The explicit data format of 3D GS, which is easy to understand and process, is friendly to downstream tasks (e.g., segmentation \cite{chen2024gaussianeditor}) and promising for the standardization study. However, also ascribe to the explicit data format and almost unrestricted ellipsoid densification (i.e., clone and split) during generation, 3D GS generally require volume memory and storage. For example, a set of 1 million 3D Gaussians is around 236 MB with 32-bit floating. Therefore, 3D GS compression emerges as an extremely important and unavoidable technology.

Theoretically, 3D GS compression can be divided into two branches: generative and traditional compression methods. For the former, the core idea is optimizing 3D GS parameters by adding extra constraint and form a more compact representation, which can release the memory and storage burden accompanying the process of GS generation. Evidence has been provided that vanilla GS has many redundant primitives \cite{lee2023compact}. One straightforward strategy is to use as few primitives as possible to realize scene representation at the expense of a certain loss of perceptual quality, such as code books used in \cite{navaneet2023compact3d, niedermayr2023compressed, fan2023lightgaussian}. Considering that the implicit data form is more compact, some methods propose embedding primitive information \cite{chen2024hac,lu2024scaffold} into the Multilayer Perceptron (MLP). Inspired by information theory, \cite{girish2023eagles} noticed that floating vectors are ineffective for compression and proposed a lightweight method to map them to integer vectors using an MLP.

For the traditional compression method, it is similar to image, video, and point cloud compression that have been widely studied in the prestige standardization working community \cite{pourazad2012hevc,pcc-mpeg}: given a pristine 3D GS, judicious serial tools are designed to post-process 3D GS data, followed by encoding the necessary data into binary stream. The corresponding decoder can recover the original 3D GS data losslessly or lossily.  However, due to the newfangled 3D representation and insufficient awareness of GS distortion characteristics and quantification, few researches explore traditional compression for 3D GS.

\begin{figure}[ht] {
\centering
    \setlength{\abovedisplayskip}{0pt}
    \setlength{\belowdisplayskip}{0pt}
\vspace{-1em}
    \includegraphics[width=0.45\textwidth]{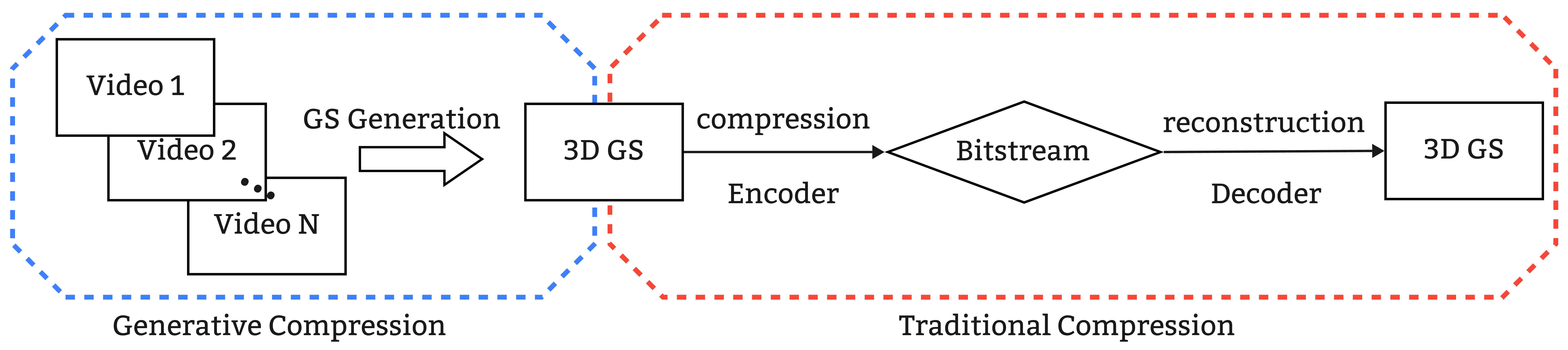}
    \caption{End-to-end GS compression system.}
    \label{fig:mov} }
    \vspace{-1.5em}
\end{figure}

Although two compression branches have different motivations, sometimes they are not conflicting and can even facilitate each other as Fig. \ref{fig:mov} shown: ideally, given Multiview Videos (MV), the generative compression method can generate 3D GS samples with compact data representation, i.e., using fewer but more effective primitive to represent this 3D scene, which can be considered as the first step of compression. Then, the traditional compression method can further compress the 3D GS into code streams for network transmission and distribution, which is the second step of compression. Considering that the generative compression method has achieved impressive results, the traditional compression method deserves commensurate attention.

For traditional compression, before developing a sophisticated and canonical compression tool, an easy-to-understand anchor resorts to well-known compression technologies and a dataset \cite{yang2020predicting} need to be constructed first, which are the two contributions of this paper. Inspired by \cite{shaovcip}, we propose an easy but effective 3D GS compression anchor, named Graph-based GS Compression (GGSC), and use different branches to compress the primitive center and attributes of GS \cite{VVM-cfp}, as shown in Fig. \ref{fig:flowchart}. GGSC first uses KDTree to split the GS into multiple subgroups, i.e., sub-GS, then the primitive centers are regarded as graph vertices to construct local graph representations, and the primitive attributes, including SH, opacity, scale, and rotation, are defined as graph signals. After Graph Fourier Transform (GFT), we clip the high-frequency components to generate multiple signal residual matrix with respect to different attributes. Considering that all the GS information is floating vectors, we adopt a simple quantization method to regularize both the primitive center and the signal residual matrix \cite{VVM-cfp}. Finally, primitive centers, which can be considered as a point cloud, are compressed by the lossless model of the Geometry-based Point Cloud Compression tool (G-PCC) \cite{GPCCTestModel} (bitrate is denoted as $\rm B1$), while the signal residual matrix is compressed by adaptive arithmetic coding (bitrate is denoted as $\rm B2$). The overall bitrate is equal to $\rm B1+B2$.

To analyze the influence of lossy compression on visual quality and facilitate the following study on GS compression, we construct a large-scale GS dataset, called GS Quality Assessment (GSQA), and perform a subjective experiment to collect subjective scores. GSQA has two parts: static and dynamic GS content. The static part consists of nine synthetic objects from Explicit\_NeRF\_QA \cite{xing2024explicit_nerf_qa} and NeRF-synthetic \cite{nerf}, and four unbounded photo-realistic scenes from Mip-NeRF360 \cite{barron2022mip}, deep blending \cite{hedman2018deep} and tanks\&temples. For dynamic GS content, we use two MV with 250 frames for each in PKU-DyMVHumans \cite{zheng2024pku} as the source. In total, we generate $9+4+250\time2 = 513$ frames of GS samples with official implementation \cite{GS}. Considering that there are two modules in GGSC can incur distortion: high-frequency clipping and quantization, we design superimposed and individual impairments. First, we use GGSC to generate five superimposed degradation levels for all GS content to study the Bjøntegaard Delta Rate-Distortion (RD) curves. Second, we generate five individual distortion levels for all the synthetic object content to study characteristics of different attribute distortions. After rendering GS into Processed Video Sequences (PVS), we generate $9\times(5+5) + (4+2)\times5 = 120$ samples for subjective experiment (250 frames from the same MV are rendered into one PVS). We validate the diversity and reliability of the GSQA, followed by a comprehensive analysis on distortion characteristics.

The remainder of this paper is organized as follows. Section \ref{sec:GGSC} details the proposed GGSC. Section \ref{sec:GSD} demonstrates the construction of the GSQA. Section \ref{sec:metric} reports the performance of state-of-the-art (SOTA) objective metrics on GSQA. Section \ref{sec:DC} analyzes the characteristics of GS distortion. Section \ref{sec:conclusion} summarizes the entire paper.

\section{Graph-based Gaussian Splatting Compression}\label{sec:GGSC}

\begin{figure}[htbp] {
    \setlength{\abovedisplayskip}{-5pt}
    \setlength{\belowdisplayskip}{-2pt}
\vspace{-1em}
\centering
     \includegraphics[width=0.35\textwidth]{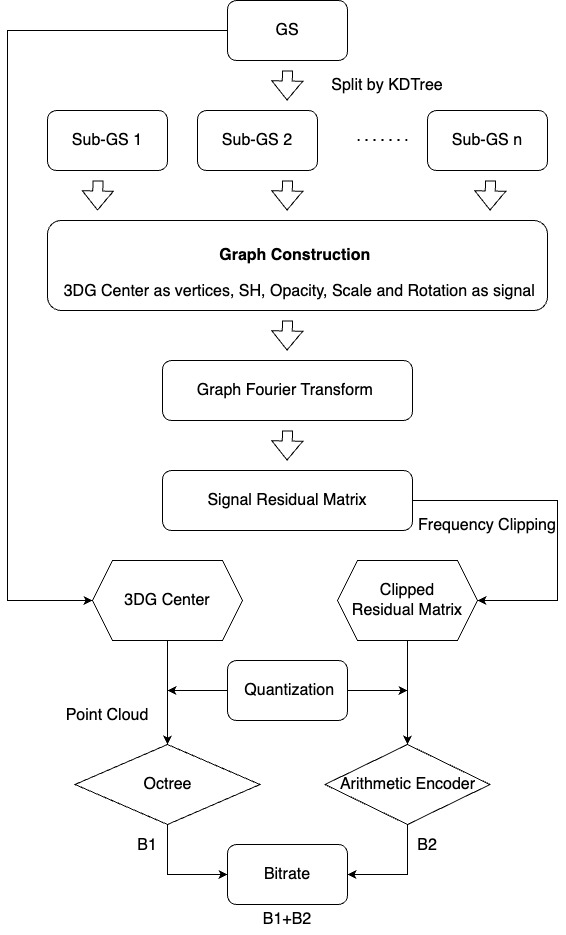}
    \caption{Flowchart of GGSC.}
    \label{fig:flowchart} }
    \vspace{-2em}
\end{figure}

In this section, we introduce the proposed GGSC in detail. Fig. \ref{fig:flowchart} illustrates the flowchart of GGSC, which consists of several main steps: GS splitting, graph construction, signal residual matrix generation, quantization, and coding.

\subsection{GS Splitting}
A GS sample can have tens of thousands to millions of primitives, to reduce calculation complexity and improve execution efficiency, we first split the whole GS into multiple sub-GS. Referring to \cite{shaovcip}, KDTree can generate more evenly split results compared to Octree, which can facilitate the following graph-related operations. Therefore, we also adopt KDTree to split GS based on primitive centers:
\begin{align}
    GS = GS_{1}^{m_1}\oplus GS_{2}^{m_2}\oplus...\oplus GS_{i}^{m_i} ... \oplus GS_{n}^{m_n},
\end{align}
$GS_{i}^{m_i}$ denotes the $i$th sub-GS that has $m_i$ primitives, $\sum_1^n m_i = |GS|$, $|GS|$ is the total number of input primitives. In this paper, we set $m_i\leq 200$.

\subsection{Graph Construction}
For each sub-GS $GS_i$, Eq. \eqref{fc: adjancy matrix} is used to construct the adjacency matrix $\mathbf{W}$ of the graph by treating primitive centers as graph vertices,
\begin{equation}\label{fc: adjancy matrix}
\mathbf{W}_{v_p^i,v_q^i}=
e^{-\frac{\|v_{p}^i-v_{q}^i\|^{2}_{2}}{\sigma^{2}}} \quad \text{if } v_{p}^i, v_{q}^i\in GS_i,
\end{equation}
$\mathbf{W} \in R ^{m_i \times m_i}$, where $\sigma$ is the variance of graph node and set as $\sigma = H^{0.5}$, $H = min(X, Y, Z)/20$ with $X = X_{max}-X_{min}$, $Y = Y_{max}-Y_{min}$, $Z = Z_{max}-Z_{min}$. The degree matrix is subsequently calculated by $\mathbf{D}= {\sf diag}(d_{1},...,d_{m_i})\in \mathbb{R}^{m_i \times m_i}$, $d_{q}=\sum_{v_q^i} \mathbf{W}_{v^i_p,v_q^i}$.
Both $\mathbf{W}$ and $\mathbf{D}$ lead to the graph Lapacian matrix, $\mathbf{L}=\mathbf{D}-\mathbf{W}$.
$\mathbf{L}$ is a difference operand in the graph, and its eigendecomposition result is $\mathbf{L} = \mathbf{A}\mathbf{\Lambda}\mathbf{A^{-1}}$, where $\mathbf{A}$ is the eigenvector matrix used as the graph transform matrix in the next step. $\mathbf{\Lambda}$ is a diagonal matrix that includes the eigenvalues of $\mathbf{L}$, indicating different frequency components of the graph.

\subsection{Residual Matrix Generation}
For a one-dimensional graph signal $f \in R^{m_i \times 1}$, the signal residual matrix after GFT is
\begin{align}
    \mathbf{C}= (\mathbf{A^{-1}} f)^{T} \in R^{1\times m_i}
\end{align}
Most coefficients in residual matrix are small and close to zero, especially on the high dimensions that corresponding to high frequency components. Similarly to the utilization of the discrete cosine transform in image compression, clipping the high-frequency components can realize impressive data compression without introducing noticeable distortion. Therefore, a clipping ratio $\alpha\in (0,1]$ is introduced to remove high-frequency part, e.g., $\mathbf{C}_{clip}= {\lfloor C \rfloor}_{\alpha \times m_i}$,
where ${\lfloor * \rfloor}_{\alpha \times m_i}$ means extract first $\alpha \times m_i$ elements to construct a new residual matrix.

For GS samples, the graph signals are SH, opacity, scale, and rotation. We calculate the residual matrix for each. Unlike other attributes that perform GFT and frequency clipping directly, for SH, we first convert it from RGB to YUV color space inspired by experience in image compression: the Luminance Component (Y) is more important for visual fidelity than chrominances (U and V), and provides a more flexible compression parameters setting.
YUV SH can be inferred from SH in RGB space directly \cite{wong2003compression}. Using the SH for the Y as an example,
\begin{footnotesize}
\begin{align}\label{fc:c1}
    C^{Y} = \int_{0}^{2\pi}\int_{0}^{\pi}P^Y(\theta, \phi)B(\theta, \phi)sin\theta d\theta d\phi,  Y = \beta_1 R + \beta_2 G + \beta_3 B,
\end{align}
\end{footnotesize}
where $C^{Y}$ is SH for the Y, $P^Y(\theta, \phi)$ is the sampled radiance value in the Y, $B(\theta, \phi)$ is the SH basis, and $\beta_1$, $\beta_2$, $\beta_3$ are weighting factors to calculate Y with RGB values that equal to 0.299, 0.587, 0.114. The weighting parameters of calculating U and V can be inferred accordingly, which are -0.169, -0.331, 0.500 and 0.500, -0.419, -0.081, respectively. Substituting $P^Y(\theta, \phi)$ into Eq. \eqref{fc:c1} yields
\begin{footnotesize}
\begin{align}
    C^{Y} = \int_{0}^{2\pi}\int_{0}^{\pi}(\beta_1 R + \beta_2 G + \beta_3 B))Bsin\theta d\theta d\phi = \beta_1 C^R + \beta_2 C^G + \beta_3 C^B,
\end{align}
\end{footnotesize}
$C^R$, $C^G$, $C^B$ are SH in RGB color space.
\subsection{Quantization and Coding}
Before coding, a uniform quantization is applied, as suggested by \cite{VVM-cfp}, that converts the floating-point values $(a_i)_{i\in{0, ..., S-1}}$ of the primitive center and the clipped residual matrix into integer values $(b_i)_{i\in{0, ..., S-1}}$ using Eq. \eqref{fc:quantization}:
\begin{equation}\label{fc:quantization}
    \setlength{\abovedisplayskip}{0.1pt}
    \setlength{\belowdisplayskip}{0.5pt}
   b_i = \lfloor\frac{(a_i - a^{min})\times(2^q-1)}{max(a^{max}-a^{min})}+\frac{1}{2}\rfloor, \forall i \in {0, ..., S-1},
\end{equation}
where $S$ is the number of floating values, $\lfloor x \rfloor$ is the floor function that takes as input a real number $x$, and gives as output the greatest integer less than or equal $x$, $q$ is the bit depth of the quantized values, and $a^{max}$, $a^{min}$ are the maximum and minimum attribute values, respectively. After quantization, we use the lossless model of G-PCC \cite{GPCCTestModel} to compress the primitive center and adaptive arithmetic coding to compress the signal residual matrix, resulting in the total bitrate of GS $\rm B = B_1 + B_2$, $\rm B_1$ and $\rm B_2$ represent the bitrate of the primitive center and the signal residual matrix.

\section{Dataset construction}\label{sec:GSD}
In this section, we demonstrate the details about the dataset construction. The GSQA consists of two parts: static and dynamic content. Two further categories are included for static content: synthetic object and unbounded photo-realistic scene.  The dynamic content is realistic captured stage scenes that targets are human figures with black background.

\subsection{Source Information}
\subsubsection{Static Content}

\par
\begin{figure}[htbp] {
\centering
    \setlength{\abovedisplayskip}{0pt}
    \setlength{\belowdisplayskip}{2pt}
\vspace{-1em}
    \includegraphics[width=0.4\textwidth]{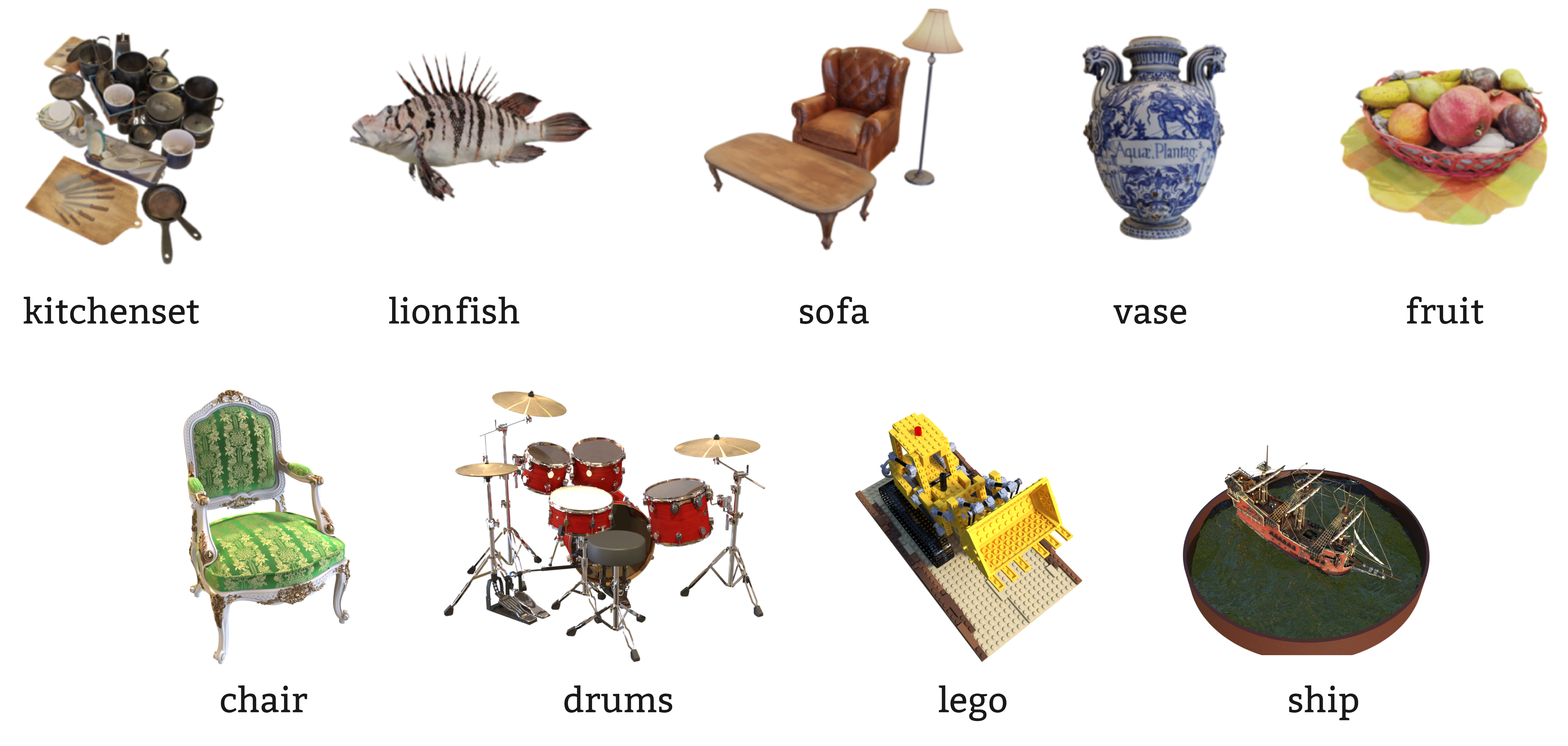}
    \caption{Snapshot of GSD source synthetic objects.}
    \label{fig:GSD-S} }
    \vspace{-1.5em}
\end{figure}

 $\bullet$ Synthetic Object: we select kitchenset, lionfish, sofa, vase, fruit from Explicit\_NeRF\_QA \cite{xing2024explicit_nerf_qa} and chair, drums, lego, ship from NeRF-synthetic \cite{nerf} as source synthetic objects of GSQA, snapshots are shown in Fig. \ref{fig:GSD-S}. Objects in Explicit\_NeRF\_QA have 130 views: 100 views for training and testing, and 30 views for validation. Objects from NeRF-synthetic have 100 views for training, 200 views for testing, and 100 views for validation. The resolution of view is $800\times 800$.



\begin{figure}[htbp] {
\centering
\vspace{-1em}
    \includegraphics[width=0.4\textwidth]{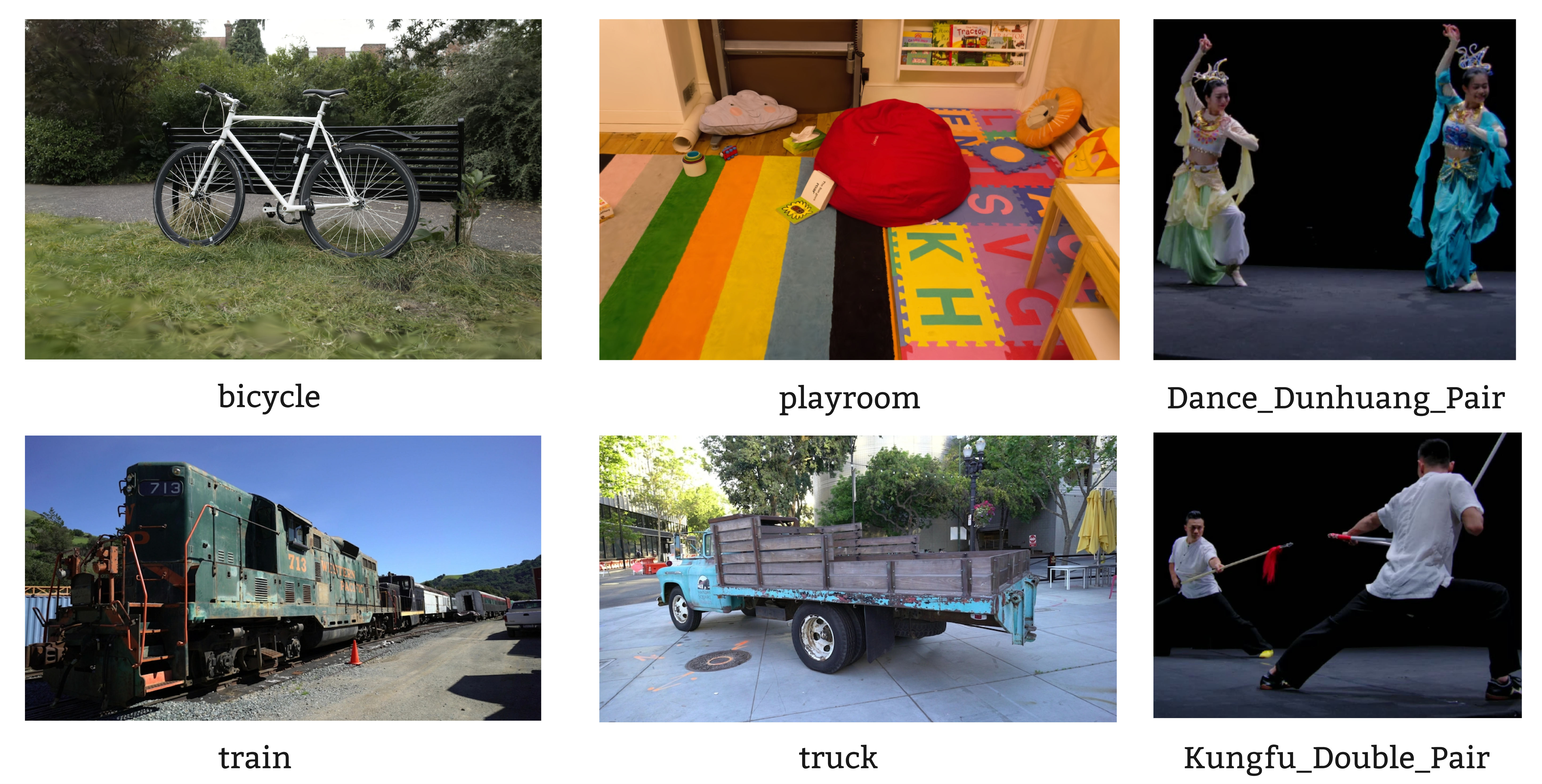}
    \caption{Snapshot of GSD source unbounded photo-realistic scenes and dynamic content.}
    \label{fig:GSD-S2} }
    \vspace{-2em}
\end{figure}

$\bullet$ Unbounded Photo-realistic Scene: there are four unbounded photo-realistic scenes come from three datasets: ``bicycle'' from Mip-NeRF360 \cite{barron2022mip}, ``playroom'' (room) from deep blending \cite{hedman2018deep}, ``train'' and ``truck'' from tanks\&temples \cite{knapitsch2017tanks}, as shown in the first and second columns of Fig. \ref{fig:GSD-S2}. ``bicycle'' has 194 views with resolution $4946\times3286$, ``room'' has 225 views with resolution $1264\times832$, ``train'' has 301 views with resolution $1959\times1090$, and ``truck'' has 251 views with resolution $1957\times1091$.

\subsubsection{Dynamic Content}
``Dance\_Dunhuang\_Pair'' (dance) and ``Kungfu\_Double\_Pair'' (kungfu) from PKU-DyMVHumans \cite{zheng2024pku} are selected as source dynamic content and shown in Fig. \ref{fig:GSD-S2} third column. Two MV provide 250 frames with 25 FPS, 60 views and each view is $1920\times1080$ resolution,  54 views are proposed for training.

\subsection{Sample Generation}
For dynamic content, we use FFMPEG \cite{ffmpeg} with FPS = 25 to extract frames from MV of the training part and merge the frames with the same timestamp as one input for GS generation. Therefore, each dynamic content can generate 250 input groups. For static content, we use the official dataset train\&testing split to generate GS samples. The default setting of the official project is used to train GS samples, and the original image resolution (without image scale) is used to avoid extra distortion. The training iteration is 30000. In all, the constructed dataset has $(9+ 4+ 2 \times 250) = 513$ reference GS samples.

\subsection{Distortion Generation}

We use GGSC to generate distorted GS samples. There are two steps in GGSC that can lead to distortion: frequency clipping and quantization. For the nine synthetic objects, we generate 5+5=10 distorted samples: five superimposed distortions and five individual distortions.  For the five superimposed distortions, we select five different bitrates considering frequency clipping and quantization simultaneously, covering the visual quality score from 0 to 10. For the five individual distortion, we fix all the parameters as lossless except one factor to reveal the influence of distortion of certain attributes on the visual quality. Each synthetic object only introduces one individual distortion with five levels: ``lionfish'': primitive center quantization (qXYZ), ``kitchenset'': SH clipping (rC), ``sofa'': opacity clipping (rO), ``fruit'': scale clipping (rS), ``vase'': rotation clipping (rR), ``lego'': SH quantization (qC), ``chair'': opacity quantization (qO), ``ship'' scale quantization (qS), and ``drums'': rotation quantization (qR). For SH, we use the same clipping ratio $\alpha$ and quantization bit depth for the three channels. For four unbounded photo-realistic scenes and two dynamic scenes, we only generate five superimposed distortion samples. For each frame of dynamic scenes, we use the same compression parameters. The detailed compression parameters are shown in the appendix. In total, we generate $9 \times 10 + 4 \times 5 + 2 \times 250 \times 5 = 2610$ distorted GS samples.

\subsection{Sample Annotation}
\subsubsection{PVS Generation}
Following a conventional approach, PVSs are generated from both the reference and the distorted GS content. These PVSs are 2D videos that depict camera paths simulating typical user motions, each frame is captured by official render software. For static content, a rotational movement is applied to capture the 3D scene \cite{kaifaICME}. For dynamic content, we fix the camera in a carefully selected view that can fully show the scene content. The resulting PVS are created with a frame rate of 30 and a duration of 20s, and subsequently are encoded by the FFMPEG library with the x265 encoder as \cite{TSMD}. For dynamic content, we render the 250 GS frames into one PVS. The PVS resolution for synthetic objects is fixed to $800\times800$, ``room'' is $1264\times832$, the other three unbounded photo-realistic scenes and two dynamic scenes are scaled to $1600\times1000$. In total, $9+4+2 = 15$ reference PVS and $9\times10 + (4+2) \times 5 = 120 $ distorted PVS are generated.
\subsubsection{Subjective Experiment}
To conduct subjective experiments, a lab experiment is implemented using a proprietary interface. Before the rating session starts, a comprehensive set of instructions is provided to the participants to ensure the clarity and consistency of the test. Then a training session is conducted to require participants to rank six carefully selected PVS, excluding any unrealistic scores \cite{TSMD}. The successful completion of the training session serves as a qualification criterion for participants to proceed to the actual rating session. The experiment follows ITU-T P.910 Recommendation \cite{11scale-rating}. A double stimulus impairment scale is used with the reference and distorted PVSs are displayed on the left and right side of the monitor at the same time, using a 11-grade scale for rating the impairments \cite{11scale-rating, yang2023tdmd}. The participants are students from universities of ages between 17 and 30.

\subsubsection{Outlier Detection}
To remove outliers and ensure the reliability of the scores collected, specific trapping PVS are intentionally included within the rating session. These trapping PVS correspond to two types: 1) extremely low quality PVS and 2) duplicated PVS. Based on the results of these trapping PVSs, it becomes possible to identify and filter out any outliers during the evaluation process. Then, an outlier detection procedure from ITU-R BT.500 Recommendation \cite{BT500} is used to remove scores generated by unreliable participants. Each PVS is guaranteed to take advantage of the remaining number of scores above 15 to compute the mean opinion scores (MOS) after the outliers are rejected.

\subsection{Dataset Validation}
In this section, we demonstrate that the constructed dataset provides samples with rich visual characteristics and accurate MOS.
\subsubsection{Diversity of Content}
\begin{figure}[htbp] {
\setlength{\abovecaptionskip}{0pt}
\setlength{\belowcaptionskip}{20pt}
\centering
\vspace{-1.2em}
    \includegraphics[width=0.4\textwidth]{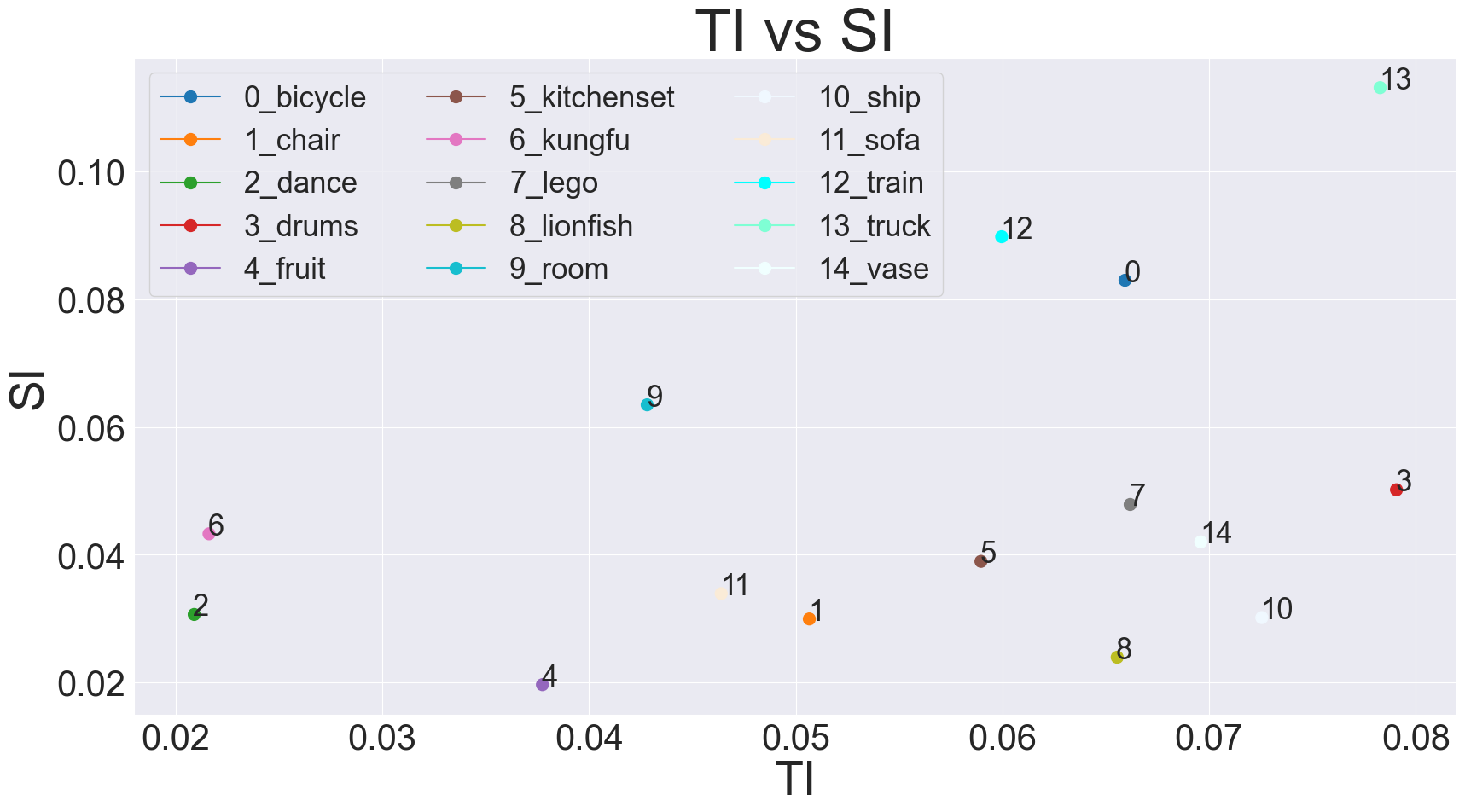}
    \caption{SI vs. TI.}
    \label{fig:SIVSTI} }
    \vspace{-3.2em}
\end{figure}

To measure the diversity of the dataset content, spatial information (SI) and temporal information (TI) are calculated based on the reference PVS used in subjective experiments. SI measures texture structures, while TI reflects motion variances. Fig. \ref{fig:SIVSTI} illustrates the SI vs TI plot. GSQA covers a wide range of SI and TI values. ``truck'' provide the largest SI, and ``drums'' show the largest TI. For the four unbounded photo-realistic scenes (i.e., Nos. 0, 9, 12, and 13), they consist of multiple contents within one scenario, resulting in generally higher SI and TI values.

\begin{figure}[ht]
\setlength{\abovecaptionskip}{2pt}
\vspace{-1.3em}
\centering
\subfigure{
\begin{minipage}[t]{0.24\textwidth}
\centering
\includegraphics[width=\textwidth]{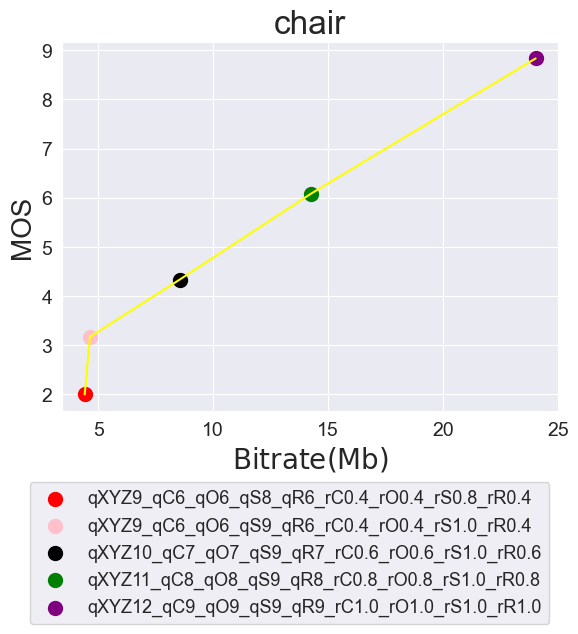}
\end{minipage}%
}%
\subfigure{
\begin{minipage}[t]{0.24\textwidth}
\centering
\includegraphics[width=\textwidth]{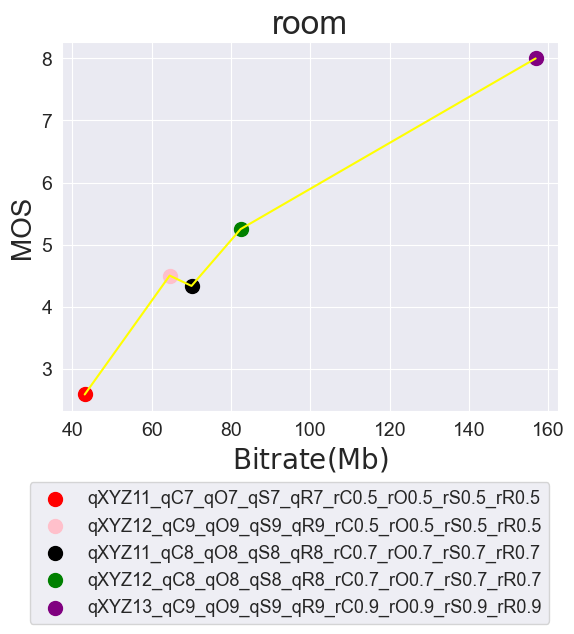}
\end{minipage}%
}%
\centering
\vspace{-1em}
\caption{MOS vs Bitrate curve.}
\vspace{-2.5em}
\label{fig:mosVSbit}
\end{figure}

\subsubsection{Accuracy of MOS}
To versify the MOS accuracy of the dataset, we examine the correlation between MOS and bitrate. For a given GS, a higher bitrate is supposed to indicate more detailed texture information, and consequently higher MOS value. Fig. \ref{fig:mosVSbit} shows the MOS vs bitrate plot for two GS. Most samples' curves in the GSQA are similar to the curve of ``chair'' that illustrate perfect monotony, as shown in Fig. \ref{fig:mosVSbit} (a). Some exceptions occur, such as for the ``room'' in Fig. \ref{fig:mosVSbit} (b). When the bitrate increases from 70 to 82 MB, the MOS drops from 4.5 to 4.3. After carefully checking the PVS, we exclude the case of inaccurate MOS and agree with the opinions of the participants. Based on the compression parameters of the pink and black bitrate points, the clipping ratio $\alpha$ that increased from 0.5 to 0.7 is less important than the quantization depths that the primitive center decreased from 12 to 11 and the attributes from 9 to 8 in terms of visual quality for this sample, revealing an improper bitrate distribution. Perfect bitrates vs MOS is achieved for the majority of GS and confirms the reliability of our subjective experiment.

\section{Performance of Objective Metric}\label{sec:metric}

We have confirmed that GSQA is a reliable dataset with rich content. Therefore, it is appropriate to evaluate the SOTA objective metrics on GSQA. The major question is whether there are existing metrics that are robust and accurate for GS quality prediction.

Two types of metrics are tested: video-based metrics (PSNR, SSIM \cite{ssim}, MSSIM \cite{mssim}, 3SSIM \cite{3ssim}, VQM \cite{vqm}, and VMAF \cite{vmaf-li2016toward}), and image-based metrics (PSNR, SSIM \cite{ssim}, and LPIPS \cite{lpips}). Video-based metrics are calculated with the MSU Video Quality Measurement Tool \cite{msu} on the PVS used for subjective experiments. The image-based metrics are calculated using the software released in \cite{GS} with the images rendered by the training cameras as input.

After mapping the objective scores to a common scale using a five-parameter logistic regression \cite{video2003final}, we report the Pearson Linear Correlation Coefficient (PLCC), Spearman Rank-order Correlation Coefficient (SRCC), and Root Mean Squared Error (RMSE) in Table. \ref{tab:correlation}. It can be observed that 3SSIM reports the best results with PLCC = 0.90, SRCC = 0.89, and RMSE = 1.15. All the correlations of video-based metrics are above 0.80 except the SRCC of SSIM which is 0.78. For image-based metrics, PSNR and SSIM show very close performance with their video-based form, indicating that the synthesized view share close distortion level with training view.
\begin{table}[h]
\small
\vspace{-1em}
\begin{tabular}{|c|c|ccc|}
\hline
\multirow{2}{*}{Type}                 & \multirow{2}{*}{Metric} & \multicolumn{3}{c|}{Correlation}                             \\ \cline{3-5}
 &       & \multicolumn{1}{c|}{PLCC} & \multicolumn{1}{c|}{SRCC} & RMSE \\ \hline
\multirow{6}{*}{Video-based   metric} & PSNR                    & \multicolumn{1}{c|}{0.82} & \multicolumn{1}{c|}{0.80} & 1.48 \\ \cline{2-5}
 & SSIM  & \multicolumn{1}{c|}{0.82} & \multicolumn{1}{c|}{0.78} & 1.48 \\ \cline{2-5}
 & MSSIM & \multicolumn{1}{c|}{0.88} & \multicolumn{1}{c|}{0.85} & 1.22 \\ \cline{2-5}
 & 3SSIM & \multicolumn{1}{c|}{0.90} & \multicolumn{1}{c|}{0.89} & 1.15 \\ \cline{2-5}
 & VQM   & \multicolumn{1}{c|}{0.83} & \multicolumn{1}{c|}{0.80} & 1.44 \\ \cline{2-5}
 & VMAF  & \multicolumn{1}{c|}{0.88} & \multicolumn{1}{c|}{0.88} & 1.23 \\ \hline
\multirow{3}{*}{Image-based   metric} & PSNR                    & \multicolumn{1}{c|}{0.82}     & \multicolumn{1}{c|}{0.77}     &    1.50  \\ \cline{2-5}
 & SSIM  & \multicolumn{1}{c|}{0.84}     & \multicolumn{1}{c|}{0.79}     &  1.42    \\ \cline{2-5}
 & LPIPS & \multicolumn{1}{c|}{0.82}     & \multicolumn{1}{c|}{0.77}     &   1.49   \\ \hline
\end{tabular}
\caption{Performance of objective metric.}
\label{tab:correlation}
\vspace{-3em}
\end{table}

\begin{figure}[ht]
\vspace{-1em}
\setlength{\abovecaptionskip}{0pt}
\setlength{\belowcaptionskip}{20pt}
\centering
\subfigure{
\begin{minipage}[t]{0.2\textwidth}
\centering
\includegraphics[width=\textwidth]{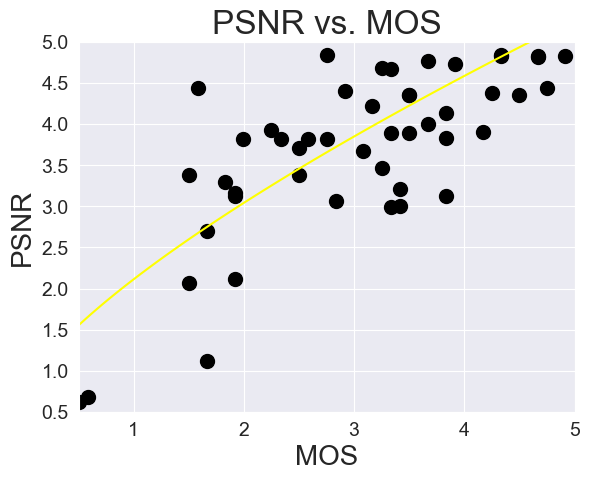}
\end{minipage}%
}%
\subfigure{
\begin{minipage}[t]{0.2\textwidth}
\centering
\includegraphics[width=\textwidth]{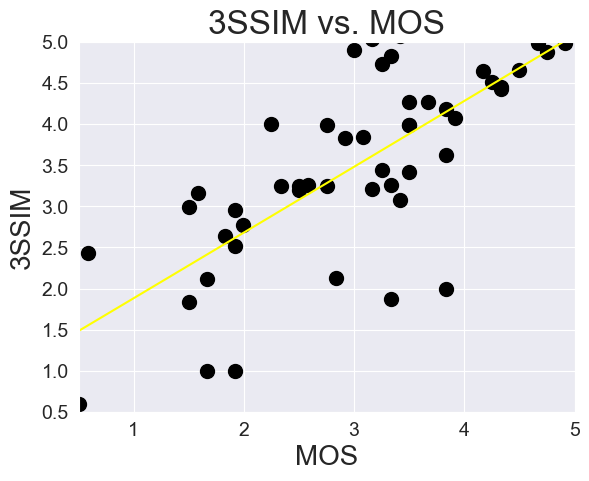}
\end{minipage}%
}
\centering
\caption{Scatter plot of metrics.}
\vspace{-4em}
\label{fig:mosVSobj}
\end{figure}

The video-based metrics report good overall correlation with GSQA, to extend the analysis of metric performance, we illustrate the scatter plots of PSNR and 3SSIM in Fig. \ref{fig:mosVSobj}. Although these metrics can realize correlations above 0.80, they all have many points away from the best-fitted logistic regression curve, represented in yellow. It indicates that better objective metrics need to be designed for GS distortion quantification.




\section{Distortion Characteristic}\label{sec:DC}
As introduced earlier, the proposed GGSC can cause distortions with the following two steps: frequency clipping and quantization. In this section, we discuss the influence of these two operations on visual quality.
\subsection{Frequency Clipping}
\begin{figure}[h]
\vspace{-1.5em}
\setlength{\abovecaptionskip}{0pt}
\setlength{\belowcaptionskip}{2pt}
\centering

\subfigure{
\begin{minipage}[t]{0.24\textwidth}
\centering
\includegraphics[width=\textwidth]{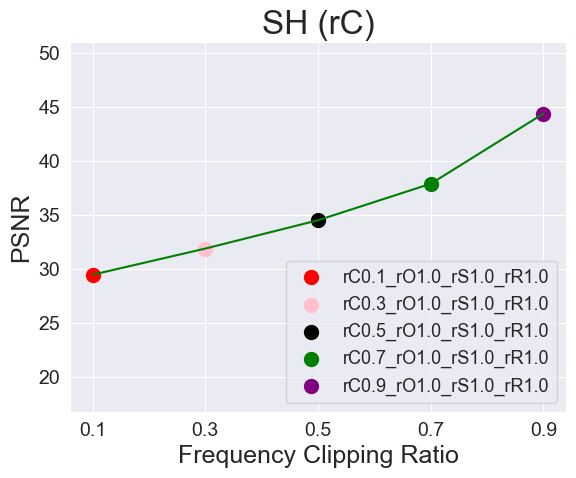}
\end{minipage}%
}%
\subfigure{
\begin{minipage}[t]{0.24\textwidth}
\centering
\includegraphics[width=\textwidth]{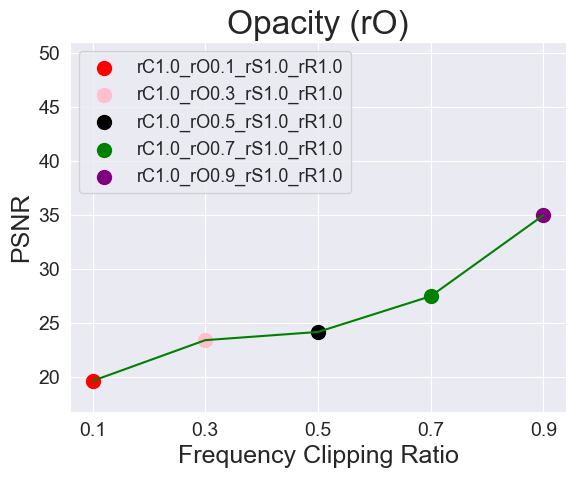}
\end{minipage}%
}\\
\subfigure{
\begin{minipage}[t]{0.24\textwidth}
\centering
\includegraphics[width=\textwidth]{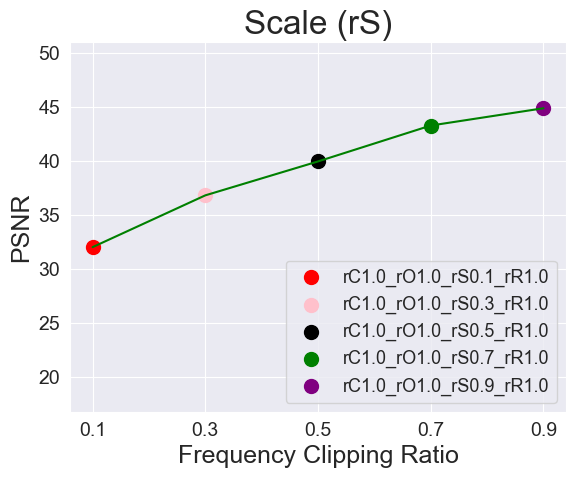}
\end{minipage}
}%
\subfigure{
\begin{minipage}[t]{0.24\textwidth}
\centering
\includegraphics[width=\textwidth]{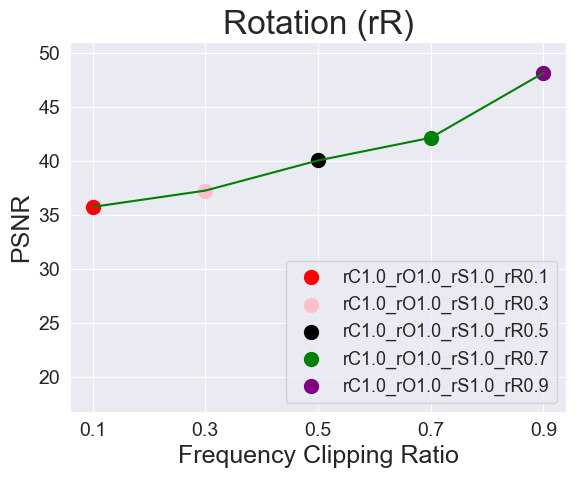}
\end{minipage}
}%
\centering
\caption{Clipping ratiob vs. Objective Metric.}
\vspace{-1.5em}
\label{fig:clip_curve}
\end{figure}

The proposed GGSC allows us to study the impact of high-frequency clipping with respect to different components of the GS attribute. The first frame of ``dance'' is selected as the study object, for SH, opacity, scale, and rotation, the clipping ratio $\alpha$ is set to 0.1, 0.3, 0.5, 0.7, and 0.9. We use PSNR to evaluate the quality of distorted GS. The curves are shown in Fig. \ref{fig:clip_curve}. It can be observed that the curves show monotonic increases with the increase of clipping ratio, which satisfies our expectation.

\begin{figure}[ht] {
\setlength{\abovecaptionskip}{0pt}
\setlength{\belowcaptionskip}{2pt}
\centering
    \includegraphics[width=0.4\textwidth]{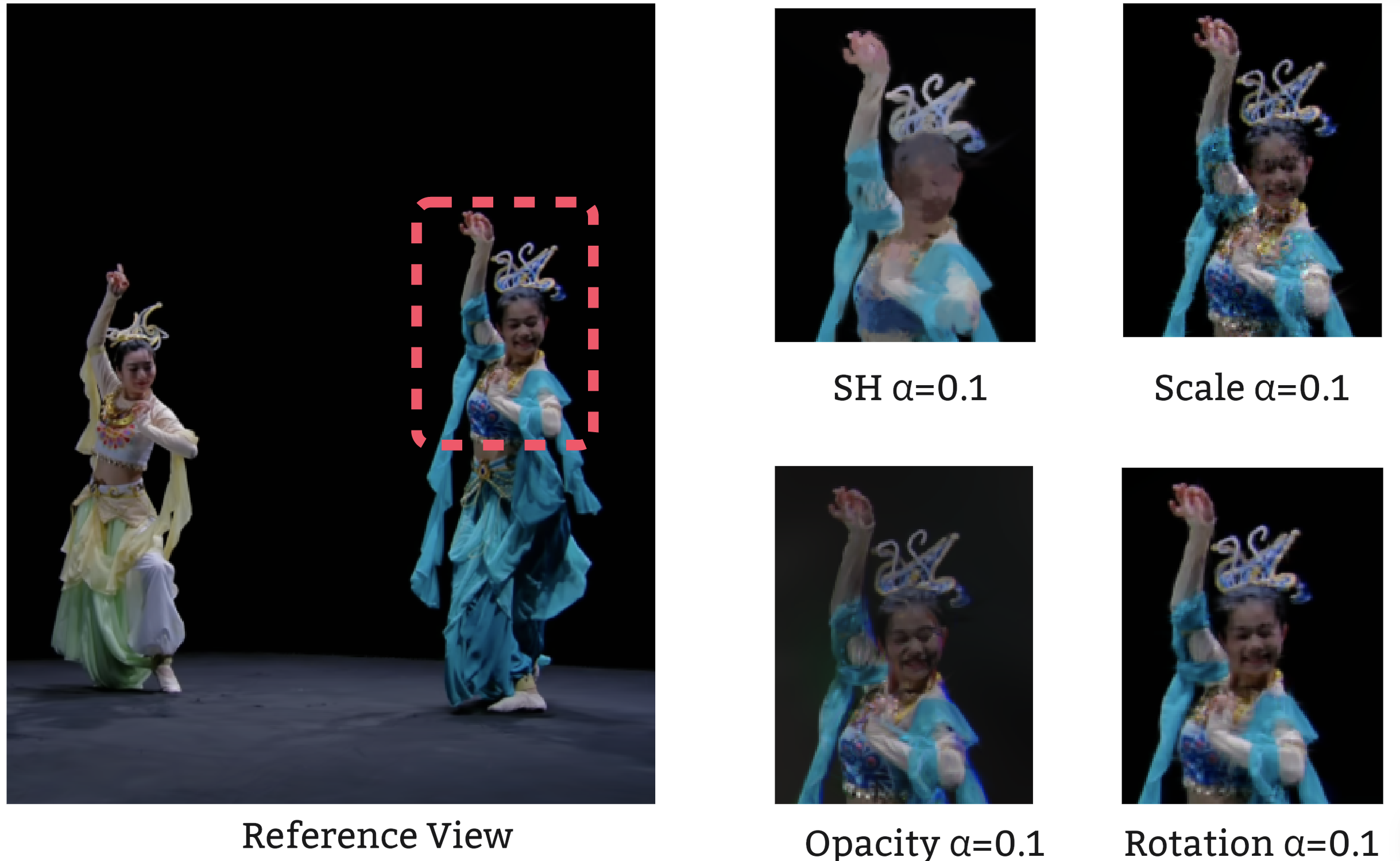}
    \caption{Snapshot of distorted samples with respect to high-frequency clipping.}
    \label{fig:clip} }
    \vspace{-1.3em}
\end{figure}

 Different attributes report different sensitivity with respect to the same clipping ratio. For example, PSNR illustrates lower quality scores for opacity (20 to 35 db) with the same level of clipping, indicating that PSNR thinks that the opacity loss is more noticeable. Fig. \ref{fig:clip} shows the rendering results with the ratio = 0.1. First, the distortions of different attributes have different visual characteristics. SH is texture blur, opacity demonstrates contrast change, scale shows high-frequency noise, while rotation reports contour blur. Second, we think that distortion of SH is more annoying than other attributes, which is a conflict with the objective metric. SH distortion is distributed mainly on the salience area, e.g., human figure, resulting in a worse perceptual quality, while opacity distortion is distributed evenly among the whole scene which is masked by the uninterested black background. A better 3D GS objective quality assessment method is urgent, and also inspires us that primitive attributes are compressible and different clipping levels can be applied for different attributes to save bitrate in real applications.

\subsection{Quantization}
\begin{figure}[ht]
\vspace{-1.5em}
\setlength{\abovecaptionskip}{0pt}
\setlength{\belowcaptionskip}{0pt}
\centering
\subfigure{
\begin{minipage}[t]{0.24\textwidth}
\centering
\includegraphics[width=\textwidth]{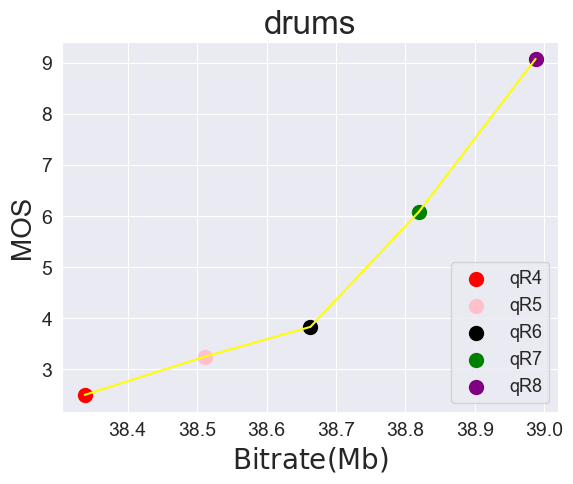}
\end{minipage}%
}%
\subfigure{
\begin{minipage}[t]{0.2\textwidth}
\centering
\includegraphics[width=\textwidth]{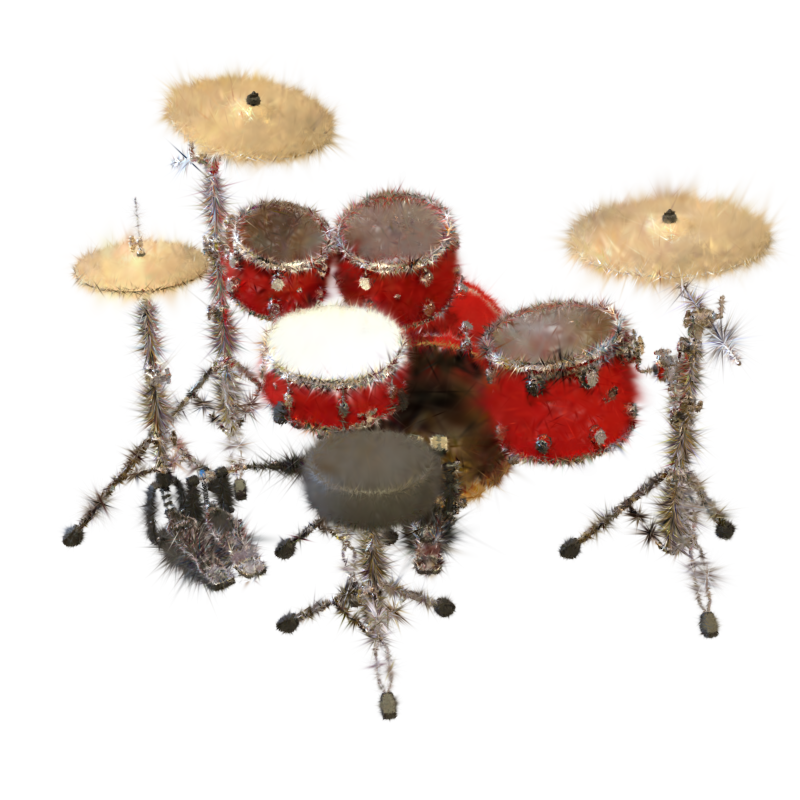}
\end{minipage}%
}\\
\subfigure{
\begin{minipage}[t]{0.2\textwidth}
\centering
\includegraphics[width=\textwidth]{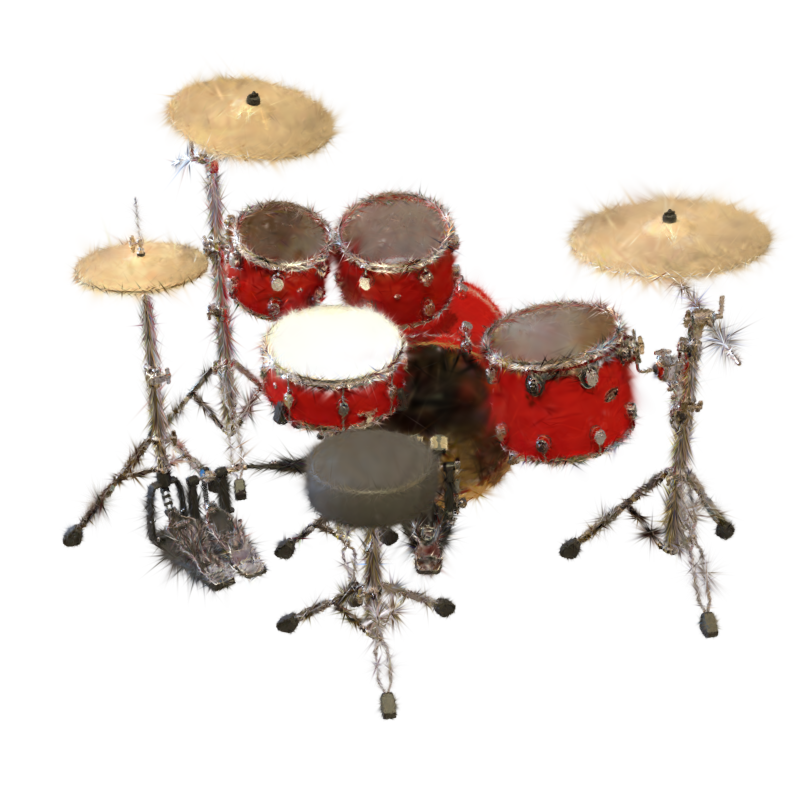}
\end{minipage}%
}%
\subfigure{
\begin{minipage}[t]{0.2\textwidth}
\centering
\includegraphics[width=\textwidth]{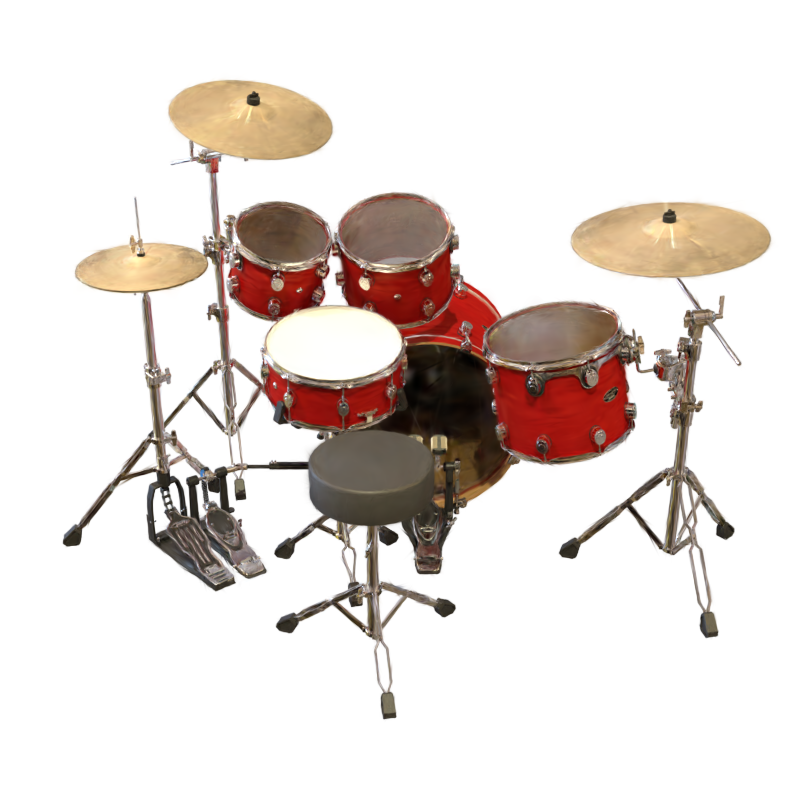}
\end{minipage}%
}%
\centering
\caption{Illustration of rotation quantization. Top left: ``drum'' MOS vs bitrate with different rotation quantization depth; Top right and bottom figures show the snapshot of ``drum'' qR = 4, 6, and 8. }
\vspace{-2em}
\label{fig:rotationQuan}
\end{figure}

Quantization can reduce the number of symbols to be coded. For the primitive center, quantization results in ellipsoid shift, overlap, and pruning; for primitive attributes, quantization reduces the richness of information. We use rotation as a representation to illustrate the influence of quantization on visual quality in Fig. \ref{fig:rotationQuan}.
It is observed that reducing the quantization depth of rotation from 8 to 4 can save bitrate from 39 to 38 MB, accompanying contour blur distortion. Considering that MOS reaches 9 when qR = 8, it demonstrates that 8 bits is enough for ``drum'' to encode primitive rotation information, and more bits are unnecessary. However, different GS samples have obvious different texture complexity, such as unbounded photo-realistic scenes have higher SI and TI, they might need larger bit depths for quality fidelity. More study is needed to decide the proper bit depths for different samples and attributes.

\section{Conclusion}\label{sec:conclusion}

To fill the gap that fewer researchers study traditional GS compression, in this paper, we propose a 3D Gaussian splatting (GS) compression anchor, called Graph-based GS compression (GGSC), and a large-scale GS Quality Assessment dataset (GSQA). First, the proposed GGSC uses two different branches to compress the GS primitive center and attributes. Followed by a KDTree split, primitive centers of each sub-GS are used to construct local graphs, serving as a basis for the following high-frequency attribute signal clipping. After quantization,  the primitive center and attribute residual matrix are compressed by G-PCC and adaptive arithmetic coding to generate the bitrate. Second, we construct GSQA to study the influence of lossy compression on human perception. GSQA consists of static and dynamic content with diverse characteristics and accurate subjective scores. Based on GSQA, we report the performance of state-of-the-art objective metrics and analyze the sensitivity of different attribute distortions in terms of human perception. The conclusions are that more robust objective metrics are needed to quantify GS distortion, and algorithms for approximate proper frequency clipping and quantization deserve considerable attention.

Among the future works, we aim to study an optimal GS compression parameter decision with respect to certain bitrate constraints \cite{liuTIP}, and develop more effective compression algorithms concerning the special data format of GS. GS data share a close format with the point cloud; therefore, we believe that study on the point cloud \cite{suTMM,GraphSIM,yang2021mped} can facilitate GS development. Besides, considering that there are some generative methods that might convert GS data into other from, e.g., implicit data, we strive to enhance the compatibility of different compression strategies.


\bibliographystyle{ACM-Reference-Format}
\bibliography{sample-base}


\begin{thebibliography}{40}


\ifx \showCODEN    \undefined \def \showCODEN     #1{\unskip}     \fi
\ifx \showDOI      \undefined \def \showDOI       #1{#1}\fi
\ifx \showISBNx    \undefined \def \showISBNx     #1{\unskip}     \fi
\ifx \showISBNxiii \undefined \def \showISBNxiii  #1{\unskip}     \fi
\ifx \showISSN     \undefined \def \showISSN      #1{\unskip}     \fi
\ifx \showLCCN     \undefined \def \showLCCN      #1{\unskip}     \fi
\ifx \shownote     \undefined \def \shownote      #1{#1}          \fi
\ifx \showarticletitle \undefined \def \showarticletitle #1{#1}   \fi
\ifx \showURL      \undefined \def \showURL       {\relax}        \fi
\providecommand\bibfield[2]{#2}
\providecommand\bibinfo[2]{#2}
\providecommand\natexlab[1]{#1}
\providecommand\showeprint[2][]{arXiv:#2}

\bibitem[Antsiferova et~al\mbox{.}(2022)]%
        {msu}
\bibfield{author}{\bibinfo{person}{Anastasia Antsiferova},
  \bibinfo{person}{Sergey Lavrushkin}, \bibinfo{person}{Maksim Smirnov},
  \bibinfo{person}{Aleksandr Gushchin}, \bibinfo{person}{Dmitriy~S. Vatolin},
  {and} \bibinfo{person}{Dmitriy Kulikov}.} \bibinfo{year}{2022}\natexlab{}.
\newblock \showarticletitle{Video compression dataset and benchmark of
  learning-based video-quality metrics}.
\newblock \bibinfo{journal}{\emph{Conf. Neural Information Processing Systems
  Datasets and Benchmarks Track}} (\bibinfo{year}{2022}).
\newblock


\bibitem[Barron et~al\mbox{.}(2022)]%
        {barron2022mip}
\bibfield{author}{\bibinfo{person}{Jonathan~T Barron}, \bibinfo{person}{Ben
  Mildenhall}, \bibinfo{person}{Dor Verbin}, \bibinfo{person}{Pratul~P
  Srinivasan}, {and} \bibinfo{person}{Peter Hedman}.}
  \bibinfo{year}{2022}\natexlab{}.
\newblock \showarticletitle{Mip-nerf 360: Unbounded anti-aliased neural
  radiance fields}. In \bibinfo{booktitle}{\emph{Proc. IEEE/CVF Conf. Computer
  Vision and Pattern Recognition}}. \bibinfo{pages}{5470--5479}.
\newblock


\bibitem[Chen et~al\mbox{.}(2024a)]%
        {chen2024gaussianeditor}
\bibfield{author}{\bibinfo{person}{Yiwen Chen}, \bibinfo{person}{Zilong Chen},
  \bibinfo{person}{Chi Zhang}, \bibinfo{person}{Feng Wang},
  \bibinfo{person}{Xiaofeng Yang}, \bibinfo{person}{Yikai Wang},
  \bibinfo{person}{Zhongang Cai}, \bibinfo{person}{Lei Yang},
  \bibinfo{person}{Huaping Liu}, {and} \bibinfo{person}{Guosheng Lin}.}
  \bibinfo{year}{2024}\natexlab{a}.
\newblock \showarticletitle{Gaussianeditor: Swift and controllable 3d editing
  with gaussian splatting}. In \bibinfo{booktitle}{\emph{Proc. the IEEE/CVF
  Conf. Computer Vision and Pattern Recognition}}.
  \bibinfo{pages}{21476--21485}.
\newblock


\bibitem[Chen et~al\mbox{.}(2024b)]%
        {chen2024hac}
\bibfield{author}{\bibinfo{person}{Yihang Chen}, \bibinfo{person}{Qianyi Wu},
  \bibinfo{person}{Jianfei Cai}, \bibinfo{person}{Mehrtash Harandi}, {and}
  \bibinfo{person}{Weiyao Lin}.} \bibinfo{year}{2024}\natexlab{b}.
\newblock \showarticletitle{HAC: Hash-grid Assisted Context for 3D Gaussian
  Splatting Compression}.
\newblock \bibinfo{journal}{\emph{arXiv preprint arXiv:2403.14530}}
  (\bibinfo{year}{2024}).
\newblock


\bibitem[Fan et~al\mbox{.}(2023)]%
        {fan2023lightgaussian}
\bibfield{author}{\bibinfo{person}{Zhiwen Fan}, \bibinfo{person}{Kevin Wang},
  \bibinfo{person}{Kairun Wen}, \bibinfo{person}{Zehao Zhu},
  \bibinfo{person}{Dejia Xu}, {and} \bibinfo{person}{Zhangyang Wang}.}
  \bibinfo{year}{2023}\natexlab{}.
\newblock \showarticletitle{Lightgaussian: Unbounded 3d gaussian compression
  with 15x reduction and 200+ fps}.
\newblock \bibinfo{journal}{\emph{arXiv preprint arXiv:2311.17245}}
  (\bibinfo{year}{2023}).
\newblock


\bibitem[FFMPEG({[n.\,d.]})]%
        {ffmpeg}
\bibfield{author}{\bibinfo{person}{FFMPEG}.}
  \bibinfo{year}{[n.\,d.]}\natexlab{}.
\newblock \showarticletitle{{A complete, cross-platform solution to record,
  convert and stream audio and video.}}
\newblock \bibinfo{journal}{\emph{\url{https://ffmpeg.org/}}}
  (\bibinfo{year}{[n.\,d.]}).
\newblock


\bibitem[Girish et~al\mbox{.}(2023)]%
        {girish2023eagles}
\bibfield{author}{\bibinfo{person}{Sharath Girish}, \bibinfo{person}{Kamal
  Gupta}, {and} \bibinfo{person}{Abhinav Shrivastava}.}
  \bibinfo{year}{2023}\natexlab{}.
\newblock \showarticletitle{Eagles: Efficient accelerated 3d gaussians with
  lightweight encodings}.
\newblock \bibinfo{journal}{\emph{arXiv preprint arXiv:2312.04564}}
  (\bibinfo{year}{2023}).
\newblock


\bibitem[Hedman et~al\mbox{.}(2018)]%
        {hedman2018deep}
\bibfield{author}{\bibinfo{person}{Peter Hedman}, \bibinfo{person}{Julien
  Philip}, \bibinfo{person}{True Price}, \bibinfo{person}{Jan-Michael Frahm},
  \bibinfo{person}{George Drettakis}, {and} \bibinfo{person}{Gabriel Brostow}.}
  \bibinfo{year}{2018}\natexlab{}.
\newblock \showarticletitle{Deep blending for free-viewpoint image-based
  rendering}.
\newblock \bibinfo{journal}{\emph{ACM Trans. Graphics}} \bibinfo{volume}{37},
  \bibinfo{number}{6} (\bibinfo{year}{2018}), \bibinfo{pages}{1--15}.
\newblock


\bibitem[ITU-R~RECOMMENDATION(2002)]%
        {BT500}
\bibfield{author}{\bibinfo{person}{BT ITU-R~RECOMMENDATION}.}
  \bibinfo{year}{2002}\natexlab{}.
\newblock \showarticletitle{Methodology for the subjective assessment of the
  quality of television pictures}.
\newblock \bibinfo{journal}{\emph{International Telecommunication Union}}
  (\bibinfo{year}{2002}).
\newblock


\bibitem[ITU-T~RECOMMENDATION(1999)]%
        {11scale-rating}
\bibfield{author}{\bibinfo{person}{P ITU-T~RECOMMENDATION}.}
  \bibinfo{year}{1999}\natexlab{}.
\newblock \showarticletitle{Subjective video quality assessment methods for
  multimedia applications}.
\newblock \bibinfo{journal}{\emph{International Telecommunication Union}}
  (\bibinfo{year}{1999}).
\newblock


\bibitem[Kerbl et~al\mbox{.}(2023)]%
        {GS}
\bibfield{author}{\bibinfo{person}{Bernhard Kerbl}, \bibinfo{person}{Georgios
  Kopanas}, \bibinfo{person}{Thomas Leimk{\"u}hler}, {and}
  \bibinfo{person}{George Drettakis}.} \bibinfo{year}{2023}\natexlab{}.
\newblock \showarticletitle{3D Gaussian Splatting for Real-Time Radiance Field
  Rendering.}
\newblock \bibinfo{journal}{\emph{ACM Trans. Graph.}} \bibinfo{volume}{42},
  \bibinfo{number}{4} (\bibinfo{year}{2023}), \bibinfo{pages}{139--1}.
\newblock


\bibitem[Knapitsch et~al\mbox{.}(2017)]%
        {knapitsch2017tanks}
\bibfield{author}{\bibinfo{person}{Arno Knapitsch}, \bibinfo{person}{Jaesik
  Park}, \bibinfo{person}{Qian-Yi Zhou}, {and} \bibinfo{person}{Vladlen
  Koltun}.} \bibinfo{year}{2017}\natexlab{}.
\newblock \showarticletitle{Tanks and temples: Benchmarking large-scale scene
  reconstruction}.
\newblock \bibinfo{journal}{\emph{ACM Trans. Graphics}} \bibinfo{volume}{36},
  \bibinfo{number}{4} (\bibinfo{year}{2017}), \bibinfo{pages}{1--13}.
\newblock


\bibitem[Lee et~al\mbox{.}(2023)]%
        {lee2023compact}
\bibfield{author}{\bibinfo{person}{Joo~Chan Lee}, \bibinfo{person}{Daniel Rho},
  \bibinfo{person}{Xiangyu Sun}, \bibinfo{person}{Jong~Hwan Ko}, {and}
  \bibinfo{person}{Eunbyung Park}.} \bibinfo{year}{2023}\natexlab{}.
\newblock \showarticletitle{Compact 3D Gaussian Representation for Radiance
  Field}.
\newblock \bibinfo{journal}{\emph{arXiv preprint arXiv:2311.13681}}
  (\bibinfo{year}{2023}).
\newblock


\bibitem[Li and Bovik(2009)]%
        {3ssim}
\bibfield{author}{\bibinfo{person}{Chaofeng Li} {and} \bibinfo{person}{Alan~C
  Bovik}.} \bibinfo{year}{2009}\natexlab{}.
\newblock \showarticletitle{Three-component weighted structural similarity
  index}.
\newblock \bibinfo{journal}{\emph{Image quality and system performance VI}}
  \bibinfo{volume}{7242} (\bibinfo{year}{2009}), \bibinfo{pages}{252--260}.
\newblock


\bibitem[Li et~al\mbox{.}(2016)]%
        {vmaf-li2016toward}
\bibfield{author}{\bibinfo{person}{Zhi Li}, \bibinfo{person}{Anne Aaron},
  \bibinfo{person}{Ioannis Katsavounidis}, \bibinfo{person}{Anush Moorthy},
  {and} \bibinfo{person}{Megha Manohara}.} \bibinfo{year}{2016}\natexlab{}.
\newblock \showarticletitle{Toward a practical perceptual video quality
  metric}.
\newblock \bibinfo{journal}{\emph{The Netflix Tech Blog}} \bibinfo{volume}{6},
  \bibinfo{number}{2} (\bibinfo{year}{2016}), \bibinfo{pages}{2}.
\newblock


\bibitem[Liu et~al\mbox{.}(2021)]%
        {liuTIP}
\bibfield{author}{\bibinfo{person}{Qi Liu}, \bibinfo{person}{Hui Yuan},
  \bibinfo{person}{Raouf Hamzaoui}, \bibinfo{person}{Honglei Su},
  \bibinfo{person}{Junhui Hou}, {and} \bibinfo{person}{Huan Yang}.}
  \bibinfo{year}{2021}\natexlab{}.
\newblock \showarticletitle{Reduced Reference Perceptual Quality Model With
  Application to Rate Control for Video-Based Point Cloud Compression}.
\newblock \bibinfo{journal}{\emph{IEEE Trans. Image Processing}}
  \bibinfo{volume}{30} (\bibinfo{year}{2021}), \bibinfo{pages}{6623--6636}.
\newblock
\urldef\tempurl%
\url{https://doi.org/10.1109/TIP.2021.3096060}
\showDOI{\tempurl}


\bibitem[Lu et~al\mbox{.}(2024)]%
        {lu2024scaffold}
\bibfield{author}{\bibinfo{person}{Tao Lu}, \bibinfo{person}{Mulin Yu},
  \bibinfo{person}{Linning Xu}, \bibinfo{person}{Yuanbo Xiangli},
  \bibinfo{person}{Limin Wang}, \bibinfo{person}{Dahua Lin}, {and}
  \bibinfo{person}{Bo Dai}.} \bibinfo{year}{2024}\natexlab{}.
\newblock \showarticletitle{Scaffold-gs: Structured 3d gaussians for
  view-adaptive rendering}. In \bibinfo{booktitle}{\emph{Proc. the IEEE/CVF
  Conf. Computer Vision and Pattern Recognition}}.
  \bibinfo{pages}{20654--20664}.
\newblock


\bibitem[Marius.Preda(2023)]%
        {GPCCTestModel}
\bibfield{author}{\bibinfo{person}{Marius.Preda}.}
  \bibinfo{year}{2023}\natexlab{}.
\newblock \showarticletitle{Test model for geometry-based solid point cloud -
  GeS TM 3.0}.
\newblock \bibinfo{journal}{\emph{Document ISO/IEC JTC 1/SC 29/WG 7, w23324}}
  (\bibinfo{year}{2023}).
\newblock


\bibitem[Mildenhall et~al\mbox{.}(2021)]%
        {nerf}
\bibfield{author}{\bibinfo{person}{Ben Mildenhall}, \bibinfo{person}{Pratul~P.
  Srinivasan}, \bibinfo{person}{Matthew Tancik}, \bibinfo{person}{Jonathan~T.
  Barron}, \bibinfo{person}{Ravi Ramamoorthi}, {and} \bibinfo{person}{Ren Ng}.}
  \bibinfo{year}{2021}\natexlab{}.
\newblock \showarticletitle{NeRF: representing scenes as neural radiance fields
  for view synthesis}.
\newblock \bibinfo{journal}{\emph{Commun. ACM}} \bibinfo{volume}{65},
  \bibinfo{number}{1} (\bibinfo{date}{dec} \bibinfo{year}{2021}),
  \bibinfo{pages}{99–106}.
\newblock
\showISSN{0001-0782}
\urldef\tempurl%
\url{https://doi.org/10.1145/3503250}
\showDOI{\tempurl}


\bibitem[Navaneet et~al\mbox{.}(2023)]%
        {navaneet2023compact3d}
\bibfield{author}{\bibinfo{person}{KL Navaneet},
  \bibinfo{person}{Kossar~Pourahmadi Meibodi}, \bibinfo{person}{Soroush~Abbasi
  Koohpayegani}, {and} \bibinfo{person}{Hamed Pirsiavash}.}
  \bibinfo{year}{2023}\natexlab{}.
\newblock \showarticletitle{Compact3d: Compressing gaussian splat radiance
  field models with vector quantization}.
\newblock \bibinfo{journal}{\emph{arXiv preprint arXiv:2311.18159}}
  (\bibinfo{year}{2023}).
\newblock


\bibitem[Niedermayr et~al\mbox{.}(2023)]%
        {niedermayr2023compressed}
\bibfield{author}{\bibinfo{person}{Simon Niedermayr}, \bibinfo{person}{Josef
  Stumpfegger}, {and} \bibinfo{person}{R{\"u}diger Westermann}.}
  \bibinfo{year}{2023}\natexlab{}.
\newblock \showarticletitle{Compressed 3d gaussian splatting for accelerated
  novel view synthesis}.
\newblock \bibinfo{journal}{\emph{arXiv preprint arXiv:2401.02436}}
  (\bibinfo{year}{2023}).
\newblock


\bibitem[Pinson and Wolf(2004)]%
        {vqm}
\bibfield{author}{\bibinfo{person}{M.H. Pinson} {and} \bibinfo{person}{S.
  Wolf}.} \bibinfo{year}{2004}\natexlab{}.
\newblock \showarticletitle{A new standardized method for objectively measuring
  video quality}.
\newblock \bibinfo{journal}{\emph{IEEE Trans. Broadcasting}}
  \bibinfo{volume}{50}, \bibinfo{number}{3} (\bibinfo{year}{2004}),
  \bibinfo{pages}{312--322}.
\newblock
\urldef\tempurl%
\url{https://doi.org/10.1109/TBC.2004.834028}
\showDOI{\tempurl}


\bibitem[Pourazad et~al\mbox{.}(2012)]%
        {pourazad2012hevc}
\bibfield{author}{\bibinfo{person}{Mahsa~T Pourazad}, \bibinfo{person}{Colin
  Doutre}, \bibinfo{person}{Maryam Azimi}, {and} \bibinfo{person}{Panos
  Nasiopoulos}.} \bibinfo{year}{2012}\natexlab{}.
\newblock \showarticletitle{HEVC: The new gold standard for video compression:
  How does HEVC compare with H. 264/AVC?}
\newblock \bibinfo{journal}{\emph{IEEE consumer electronics magazine}}
  \bibinfo{volume}{1}, \bibinfo{number}{3} (\bibinfo{year}{2012}),
  \bibinfo{pages}{36--46}.
\newblock


\bibitem[Schwarz et~al\mbox{.}(2019)]%
        {pcc-mpeg}
\bibfield{author}{\bibinfo{person}{Sebastian Schwarz}, \bibinfo{person}{Marius
  Preda}, \bibinfo{person}{Vittorio Baroncini}, \bibinfo{person}{Madhukar
  Budagavi}, \bibinfo{person}{Pablo Cesar}, \bibinfo{person}{Philip~A. Chou},
  \bibinfo{person}{Robert~A. Cohen}, \bibinfo{person}{Maja Krivokuća},
  \bibinfo{person}{Sébastien Lasserre}, \bibinfo{person}{Zhu Li},
  \bibinfo{person}{Joan Llach}, \bibinfo{person}{Khaled Mammou},
  \bibinfo{person}{Rufael Mekuria}, \bibinfo{person}{Ohji Nakagami},
  \bibinfo{person}{Ernestasia Siahaan}, \bibinfo{person}{Ali Tabatabai},
  \bibinfo{person}{Alexis~M. Tourapis}, {and} \bibinfo{person}{Vladyslav
  Zakharchenko}.} \bibinfo{year}{2019}\natexlab{}.
\newblock \showarticletitle{Emerging MPEG Standards for Point Cloud
  Compression}.
\newblock \bibinfo{journal}{\emph{IEEE Jour. Emerging and Selected Topics in
  Circuits and Systems}} \bibinfo{volume}{9}, \bibinfo{number}{1}
  (\bibinfo{year}{2019}), \bibinfo{pages}{133--148}.
\newblock


\bibitem[Shao et~al\mbox{.}(2017)]%
        {shaovcip}
\bibfield{author}{\bibinfo{person}{Yiting Shao}, \bibinfo{person}{Zhaobin
  Zhang}, \bibinfo{person}{Zhu Li}, \bibinfo{person}{Kui Fan}, {and}
  \bibinfo{person}{Ge Li}.} \bibinfo{year}{2017}\natexlab{}.
\newblock \showarticletitle{Attribute compression of 3D point clouds using
  Laplacian sparsity optimized graph transform}. In
  \bibinfo{booktitle}{\emph{IEEE Visual Communications and Image Processing}}.
  \bibinfo{pages}{1--4}.
\newblock
\urldef\tempurl%
\url{https://doi.org/10.1109/VCIP.2017.8305131}
\showDOI{\tempurl}


\bibitem[VQEG({[n.\,d.]})]%
        {video2003final}
\bibfield{author}{\bibinfo{person}{VQEG}.} \bibinfo{year}{[n.\,d.]}\natexlab{}.
\newblock \showarticletitle{Final report from the Video Quality Experts Group
  on the validation of objective models of video quality assessment}.
\newblock
  \bibinfo{journal}{\emph{http://www.its.bldrdoc.gov/vqeg/vqeg-home.aspx}}
  (\bibinfo{year}{[n.\,d.]}).
\newblock


\bibitem[VVM(2023)]%
        {VVM-cfp}
\bibfield{author}{\bibinfo{person}{VVM}.} \bibinfo{year}{2023}\natexlab{}.
\newblock \showarticletitle{Call for proposals on static polygonal mesh
  coding}.
\newblock \bibinfo{journal}{\emph{Alliance for Open Media, document:
  VVM-2023-004o-v2}} (\bibinfo{year}{2023}).
\newblock


\bibitem[Wang et~al\mbox{.}(2004)]%
        {ssim}
\bibfield{author}{\bibinfo{person}{Zhou Wang}, \bibinfo{person}{A.C. Bovik},
  \bibinfo{person}{H.R. Sheikh}, {and} \bibinfo{person}{E.P. Simoncelli}.}
  \bibinfo{year}{2004}\natexlab{}.
\newblock \showarticletitle{Image quality assessment: from error visibility to
  structural similarity}.
\newblock \bibinfo{journal}{\emph{IEEE Trans. Image Processing}}
  \bibinfo{volume}{13}, \bibinfo{number}{4} (\bibinfo{year}{2004}),
  \bibinfo{pages}{600--612}.
\newblock
\urldef\tempurl%
\url{https://doi.org/10.1109/TIP.2003.819861}
\showDOI{\tempurl}


\bibitem[Wang et~al\mbox{.}(2003)]%
        {mssim}
\bibfield{author}{\bibinfo{person}{Zhou Wang}, \bibinfo{person}{Eero~P
  Simoncelli}, {and} \bibinfo{person}{Alan~C Bovik}.}
  \bibinfo{year}{2003}\natexlab{}.
\newblock \showarticletitle{Multiscale structural similarity for image quality
  assessment}.
\newblock \bibinfo{journal}{\emph{IEEE Asilomar Conf. Signals, Systems \&
  Computers}}  \bibinfo{volume}{2} (\bibinfo{year}{2003}),
  \bibinfo{pages}{1398--1402}.
\newblock


\bibitem[Wong and Leung(2003)]%
        {wong2003compression}
\bibfield{author}{\bibinfo{person}{Tien-Tsin Wong} {and}
  \bibinfo{person}{Chi-Sing Leung}.} \bibinfo{year}{2003}\natexlab{}.
\newblock \showarticletitle{Compression of illumination-adjustable images}.
\newblock \bibinfo{journal}{\emph{IEEE Trans. Circuits and Systems for Video
  Technology}} \bibinfo{volume}{13}, \bibinfo{number}{11}
  (\bibinfo{year}{2003}), \bibinfo{pages}{1107--1118}.
\newblock


\bibitem[Xing et~al\mbox{.}(2024)]%
        {xing2024explicit_nerf_qa}
\bibfield{author}{\bibinfo{person}{Yuke Xing}, \bibinfo{person}{Qi Yang},
  \bibinfo{person}{Kaifa Yang}, \bibinfo{person}{Yilin Xu}, {and}
  \bibinfo{person}{Zhu Li}.} \bibinfo{year}{2024}\natexlab{}.
\newblock \showarticletitle{Explicit\_NeRF\_QA: A Quality Assessment Database
  for Explicit NeRF Model Compression}.
\newblock \bibinfo{journal}{\emph{arXiv preprint arXiv:2407.08165}}
  (\bibinfo{year}{2024}).
\newblock


\bibitem[Yang et~al\mbox{.}(2023c)]%
        {kaifaICME}
\bibfield{author}{\bibinfo{person}{Kaifa Yang}, \bibinfo{person}{Qi Yang},
  \bibinfo{person}{Joel Jung}, \bibinfo{person}{Yiling Xu},
  \bibinfo{person}{Xiaozhong Xu}, {and} \bibinfo{person}{Shan Liu}.}
  \bibinfo{year}{2023}\natexlab{c}.
\newblock \showarticletitle{Exploring the Influence of View and Camera Path
  Selection for Dynamic Mesh Quality Assessment}. In
  \bibinfo{booktitle}{\emph{2023 IEEE Int. Conf. Multimedia and Expo}}.
  \bibinfo{pages}{2489--2494}.
\newblock
\urldef\tempurl%
\url{https://doi.org/10.1109/ICME55011.2023.00424}
\showDOI{\tempurl}


\bibitem[Yang et~al\mbox{.}(2020)]%
        {yang2020predicting}
\bibfield{author}{\bibinfo{person}{Qi Yang}, \bibinfo{person}{Hao Chen},
  \bibinfo{person}{Zhan Ma}, \bibinfo{person}{Yiling Xu},
  \bibinfo{person}{Rongjun Tang}, {and} \bibinfo{person}{Jun Sun}.}
  \bibinfo{year}{2020}\natexlab{}.
\newblock \showarticletitle{Predicting the perceptual quality of point cloud: A
  3D-to-2D projection-based exploration}.
\newblock \bibinfo{journal}{\emph{IEEE Trans. Multimedia}}
  \bibinfo{volume}{23} (\bibinfo{year}{2020}), \bibinfo{pages}{3877--3891}.
\newblock


\bibitem[Yang et~al\mbox{.}(2023a)]%
        {yang2023tdmd}
\bibfield{author}{\bibinfo{person}{Qi Yang}, \bibinfo{person}{Jo{\"e}l Jung},
  \bibinfo{person}{Timon Deschamps}, \bibinfo{person}{Xiaozhong Xu}, {and}
  \bibinfo{person}{Shan Liu}.} \bibinfo{year}{2023}\natexlab{a}.
\newblock \showarticletitle{Tdmd: A database for dynamic color mesh subjective
  and objective quality explorations}.
\newblock \bibinfo{journal}{\emph{arXiv preprint arXiv:2308.01499}}
  (\bibinfo{year}{2023}).
\newblock


\bibitem[Yang et~al\mbox{.}(2023b)]%
        {TSMD}
\bibfield{author}{\bibinfo{person}{Qi Yang}, \bibinfo{person}{Joel Jung},
  \bibinfo{person}{Haiqiang Wang}, \bibinfo{person}{Xiaozhong Xu}, {and}
  \bibinfo{person}{Shan Liu}.} \bibinfo{year}{2023}\natexlab{b}.
\newblock \showarticletitle{TSMD: A Database for Static Color Mesh Quality
  Assessment Study}. In \bibinfo{booktitle}{\emph{IEEE Int. Conf. Visual
  Communications and Image Processing}}. \bibinfo{pages}{1--5}.
\newblock
\urldef\tempurl%
\url{https://doi.org/10.1109/VCIP59821.2023.10402660}
\showDOI{\tempurl}


\bibitem[Yang et~al\mbox{.}(2022a)]%
        {GraphSIM}
\bibfield{author}{\bibinfo{person}{Qi Yang}, \bibinfo{person}{Zhan Ma},
  \bibinfo{person}{Yiling Xu}, \bibinfo{person}{Zhu Li}, {and}
  \bibinfo{person}{Jun Sun}.} \bibinfo{year}{2022}\natexlab{a}.
\newblock \showarticletitle{Inferring Point Cloud Quality via Graph
  Similarity}.
\newblock \bibinfo{journal}{\emph{IEEE Trans. Pattern Analysis and Machine
  Intelligence}} \bibinfo{volume}{44}, \bibinfo{number}{6}
  (\bibinfo{year}{2022}), \bibinfo{pages}{3015--3029}.
\newblock
\urldef\tempurl%
\url{https://doi.org/10.1109/TPAMI.2020.3047083}
\showDOI{\tempurl}


\bibitem[Yang et~al\mbox{.}(2022b)]%
        {yang2021mped}
\bibfield{author}{\bibinfo{person}{Qi Yang}, \bibinfo{person}{Yujie Zhang},
  \bibinfo{person}{Siheng Chen}, \bibinfo{person}{Yiling Xu},
  \bibinfo{person}{Jun Sun}, {and} \bibinfo{person}{Zhan Ma}.}
  \bibinfo{year}{2022}\natexlab{b}.
\newblock \showarticletitle{MPED: Quantifying Point Cloud Distortion Based on
  Multiscale Potential Energy Discrepancy}.
\newblock \bibinfo{journal}{\emph{IEEE Trans. Pattern Analysis and Machine
  Intelligence}} (\bibinfo{year}{2022}), \bibinfo{pages}{1--18}.
\newblock
\urldef\tempurl%
\url{https://doi.org/10.1109/TPAMI.2022.3213831}
\showDOI{\tempurl}


\bibitem[Zhang et~al\mbox{.}(2018)]%
        {lpips}
\bibfield{author}{\bibinfo{person}{Richard Zhang}, \bibinfo{person}{Phillip
  Isola}, \bibinfo{person}{Alexei~A Efros}, \bibinfo{person}{Eli Shechtman},
  {and} \bibinfo{person}{Oliver Wang}.} \bibinfo{year}{2018}\natexlab{}.
\newblock \showarticletitle{The unreasonable effectiveness of deep features as
  a perceptual metric}. In \bibinfo{booktitle}{\emph{Proc. the IEEE/CVF Conf.
  Computer Vision and Pattern Recognition}}. \bibinfo{pages}{586--595}.
\newblock


\bibitem[Zheng et~al\mbox{.}(2024)]%
        {zheng2024pku}
\bibfield{author}{\bibinfo{person}{Xiaoyun Zheng}, \bibinfo{person}{Liwei
  Liao}, \bibinfo{person}{Xufeng Li}, \bibinfo{person}{Jianbo Jiao},
  \bibinfo{person}{Rongjie Wang}, \bibinfo{person}{Feng Gao},
  \bibinfo{person}{Shiqi Wang}, {and} \bibinfo{person}{Ronggang Wang}.}
  \bibinfo{year}{2024}\natexlab{}.
\newblock \showarticletitle{PKU-DyMVHumans: A Multi-View Video Benchmark for
  High-Fidelity Dynamic Human Modeling}.
\newblock \bibinfo{journal}{\emph{arXiv preprint arXiv:2403.16080}}
  (\bibinfo{year}{2024}).
\newblock


\bibitem[Zhu et~al\mbox{.}(2024)]%
        {suTMM}
\bibfield{author}{\bibinfo{person}{Linxia Zhu}, \bibinfo{person}{Jun Cheng},
  \bibinfo{person}{Xu Wang}, \bibinfo{person}{Honglei Su},
  \bibinfo{person}{Huan Yang}, \bibinfo{person}{Hui Yuan}, {and}
  \bibinfo{person}{Jari Korhonen}.} \bibinfo{year}{2024}\natexlab{}.
\newblock \showarticletitle{3DTA: No-Reference 3D Point Cloud Quality
  Assessment with Twin Attention}.
\newblock \bibinfo{journal}{\emph{IEEE Trans. Multimedia}}
  (\bibinfo{year}{2024}), \bibinfo{pages}{1--14}.
\newblock
\urldef\tempurl%
\url{https://doi.org/10.1109/TMM.2024.3407698}
\showDOI{\tempurl}


\end{thebibliography}

\end{sloppypar}
\end{document}